\documentclass[11pt]{article}

\PassOptionsToPackage{table}{xcolor}
\usepackage[preprint]{acl}

\usepackage{times}
\usepackage{latexsym}
\usepackage[T1]{fontenc}
\usepackage[utf8]{inputenc}
\usepackage{microtype}
\usepackage{graphicx}

\usepackage{amsmath}
\usepackage{amsfonts}
\usepackage{nicefrac}

\usepackage{algorithm}
\usepackage{algorithmic}

\usepackage{booktabs}
\usepackage{multirow}
\usepackage{makecell}
\usepackage{enumitem}

\definecolor{adamtpblue}{RGB}{222, 235, 247}

\title{AdaMTP: An Adaptive Training Paradigm for Multi-Token Prediction}

\author{
  \textbf{Ziqiang Cui\textsuperscript{1}}\thanks{Email: \href{mailto:ziqiang.cui@my.cityu.edu.hk}{ziqiang.cui@my.cityu.edu.hk}},
  \textbf{Han Shi\textsuperscript{2}}\thanks{Corresponding authors.},
  \textbf{Bowei He\textsuperscript{3,4}},
  \textbf{Yu Pan\textsuperscript{2}},
  \textbf{Peiyang Liu\textsuperscript{5}},
  \textbf{Shengyin Sun\textsuperscript{1}},
\\
  \textbf{Yankai Chen\textsuperscript{3,4}},
  \textbf{Haoli Bai\textsuperscript{2}},
  \textbf{Yichun Yin\textsuperscript{2}},
  \textbf{Xue Liu\textsuperscript{3,4}},
  \textbf{Chen Ma\textsuperscript{1}}\footnotemark[2]
\\
\\
  \textsuperscript{1}City University of Hong Kong,
  \textsuperscript{2}Huawei Technologies
\\
  \textsuperscript{3}Mohamed bin Zayed University of Artificial Intelligence
\\
  \textsuperscript{4}McGill University,
  \textsuperscript{5}Peking University
}

\begin{document}

\maketitle

\begin{abstract}
Multi-Token Prediction (MTP) has emerged as an effective paradigm that augments a shared Large Language Model backbone with auxiliary heads, training the model to predict several future tokens in parallel to enrich its supervision signal and accelerate inference. However, existing training frameworks adopt a rigid, fixed-length prediction horizon, disregarding the highly non-uniform information density of natural language and code. Forcing the auxiliary heads to predict across high-entropy semantic boundaries injects noisy, conflicting training signals; because these heads share the backbone's latent representations, the resulting gradients backpropagate and interfere with the model's core capabilities. We propose \textbf{AdaMTP}, an adaptive training paradigm that dynamically aligns the prediction horizon with the intrinsic predictability of the sequence. At its core, an entropy-based segmentation algorithm leverages the base model to detect sudden surges in uncertainty as semantic boundaries, partitioning sequences into variable-length groups. Each token is assigned an adaptive prediction depth, and a dynamically masked MTP objective suppresses the loss for predictions that cross these boundaries, attenuating the noisy gradients that degrade the backbone. Across mathematical reasoning, code generation, and general benchmarks on three backbones (Llama-3.1-8B, Qwen-2.5-7B, Gemma-3-12B), AdaMTP consistently outperforms standard MTP in both task performance and inference speedup.

\end{abstract}

\section{Introduction}
Large Language Models (LLMs) have achieved remarkable success across a wide range of natural language processing, mathematical reasoning, and code generation tasks. Currently, these models are predominantly trained using the standard Next-Token Prediction (NTP) objective, where models are trained to predict the immediate next token given the preceding context. Despite their robust generative capabilities, the strictly autoregressive nature of NTP imposes a fundamental bottleneck during inference. Generating tokens one by one leads to high latency and computational inefficiency, especially for
long-form generation and interactive applications.

To mitigate this inference bottleneck, Multi-Token Prediction (MTP) has recently emerged as a highly effective paradigm. By augmenting a shared LLM backbone with multiple auxiliary output heads, MTP generalizes the standard training objective to predict several future tokens simultaneously. During training, this multi-token objective provides richer supervision signals, equipping the model with enhanced long-term planning capabilities. During inference, MTP circumvents the strict autoregressive bottleneck by generating multiple tokens per step, thereby accelerating decoding speed.

\begin{figure}[t]
\setlength{\abovecaptionskip}{-0.005mm} 
\setlength{\belowcaptionskip}{-6mm} 
  \centering
  \includegraphics[width=\columnwidth]{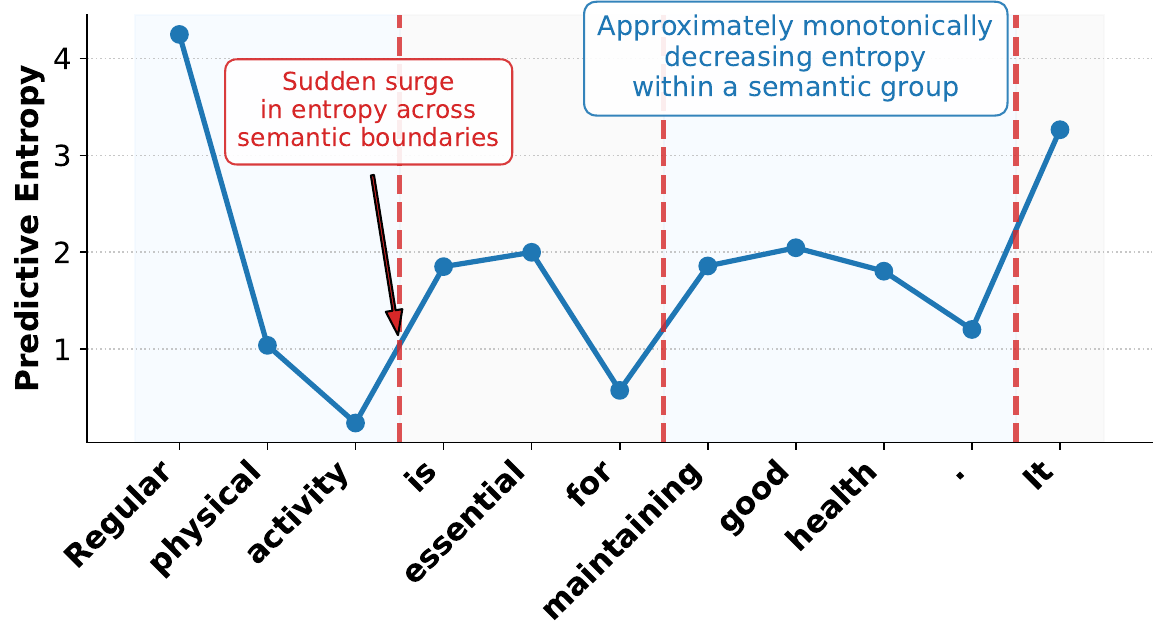}
\caption{Token-level predictive entropy of the LLM over an example sequence. Within a cohesive semantic group, the entropy follows an \textit{approximately} monotonically decreasing trend as the local context becomes increasingly constrained; at a semantic boundary, the entropy exhibits a sudden surge. AdaMTP exploits these entropy surges to partition the sequence into variable-length groups.}
   \label{fig:intro}
\end{figure}

However, existing MTP frameworks enforce a rigid, fixed-length prediction horizon for every token, overlooking the varying predictability of the underlying context. Natural language and code exhibit highly non-uniform information density. Within contiguous chunks such as common phrases or local code blocks, the sequence is highly predictable; notably, we observe that predictive uncertainty (entropy) approximately follows a monotonically decreasing trend in these regions (as illustrated in Figure~\ref{fig:intro}). In contrast, transitions between different conceptual ideas or syntactic structures are characterized by sudden spikes in uncertainty. Forcing the auxiliary heads to predict a fixed number of tokens across these high-entropy boundaries inherently injects noisy, conflicting training signals. Because the auxiliary heads and the main language modeling head share the same latent representations, these noisy gradients backpropagate and cause severe representation interference, ultimately degrading the base model's core capabilities.

Motivated by these limitations, we propose \textbf{AdaMTP} (Adaptive Multi-Token Prediction), a novel framework that dynamically aligns the multi-token prediction horizon with the intrinsic predictability of the sequence. At the core of AdaMTP is an entropy-based data segmentation algorithm. By using the base LLM to estimate token-level predictive entropy, we identify sudden surges in uncertainty---which disrupt the aforementioned approximately monotonic decrease---as semantic boundaries, partitioning the sequence into cohesive, variable-length groups. Based on these partitions, we assign each token an \textit{adaptive prediction depth}, defined as the distance to the end of its current group (or the subsequent group for boundary tokens). To incorporate this into training, we introduce a dynamically masked MTP loss: for any given token, the loss for future predictions that exceed its adaptive depth is masked out. This crucial design prevents the auxiliary heads from forcibly predicting across unpredictable boundaries, thereby attenuating the noisy gradients that degrade the model's core capabilities. 
At inference time, AdaMTP provides two decoding modes. By default, it adopts the same fixed-horizon scheme as standard MTP, yet achieves faster inference; alternatively, an adaptive-horizon mode prunes low-confidence branches via the real-time entropy signal to cut verification cost---an efficiency edge that becomes pronounced under large-batch serving.

We comprehensively evaluate AdaMTP on diverse benchmarks spanning mathematical reasoning, code generation, and general language proficiency. Using three representative base models---Llama-3.1 (8B), Qwen-2.5 (7B), and Gemma-3 (12B)---our extensive experiments demonstrate that AdaMTP improves on both task performance and inference efficiency. In terms of quality, it mitigates the representation interference inherent in standard MTP and consistently surpasses both the NTP and fixed-horizon MTP baselines in average score. In terms of efficiency, it retains and further strengthens the self-speculative acceleration of MTP, delivering substantial speedups over NTP while also decoding faster than standard MTP. Ultimately, these results indicate that adapting the MTP objective to local predictability is a more reliable way to retrofit pretrained LLMs with efficient multi-token generation.

In summary, our main contributions are as follows:
\begin{itemize}[leftmargin=1em]
    \item We identify and empirically corroborate a key limitation of existing MTP training: uniformly predicting a fixed number of future tokens can adversely affect the pretrained backbone, which we attribute to the noisy supervision forced across high-entropy linguistic boundaries.
    \item We propose AdaMTP, an adaptive training paradigm that uses entropy-based data segmentation to assign each token an adaptive prediction depth, together with a dynamically masked MTP objective that suppresses training signals beyond this depth to curb the resulting noisy gradients.
    \item Extensive experiments across three backbones and eight benchmarks show that AdaMTP consistently outperforms both NTP and standard MTP in task performance while decoding faster---delivering substantial speedups over NTP and surpassing the inference speed of standard MTP.
\end{itemize}

\section{Preliminaries}
\begin{figure*}[t]
  \centering
  \includegraphics[width=0.99\textwidth]{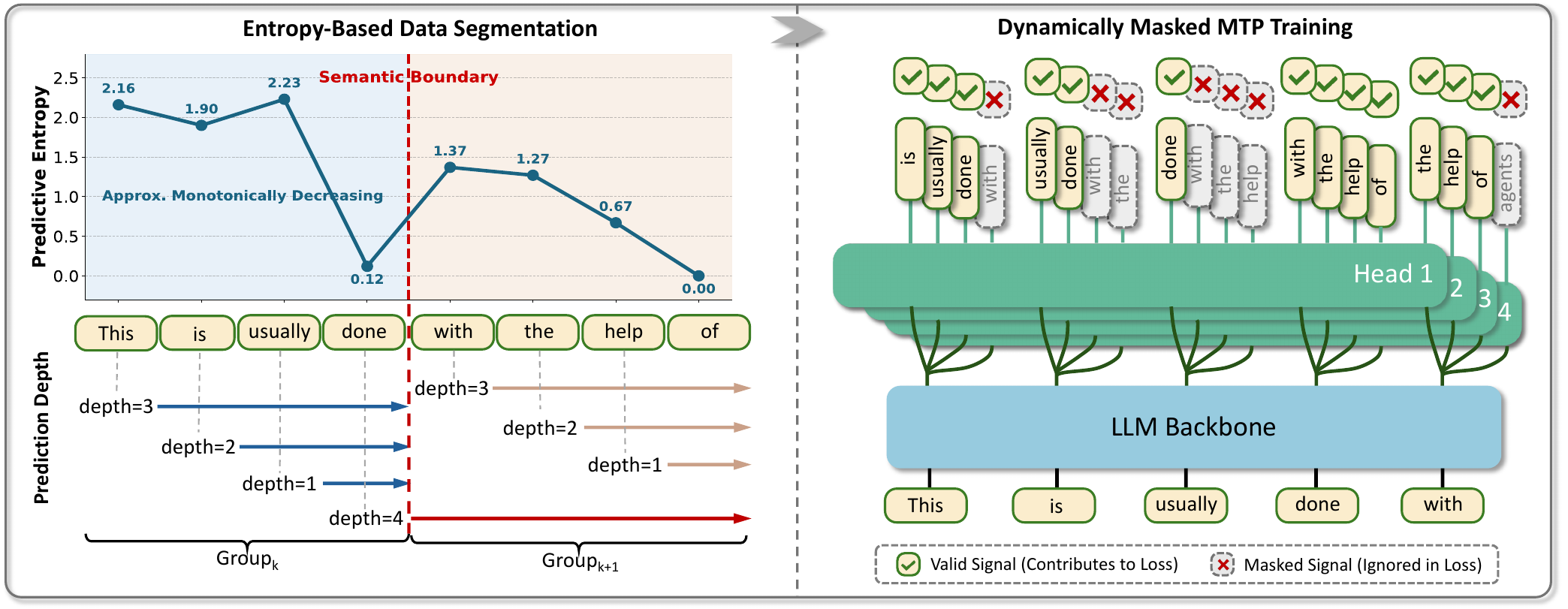}
\caption{Overview of AdaMTP training paradigm. \textbf{(Left) Entropy-Based Data Segmentation:} using the base model's next-token entropy, the sequence is split into variable-length groups at points of sudden entropy surge, and each token is assigned an adaptive prediction depth. \textbf{(Right) Dynamically Masked MTP Training:} the shared LLM backbone feeds $n$ parallel heads that predict future tokens; losses for predictions falling within each token's adaptive depth are retained as valid signal, while those crossing group boundaries are masked out and ignored.}
   \label{fig:method}
\end{figure*}

\subsection{Next-Token Prediction (NTP)}
The standard training paradigm for LLMs relies on NTP task. Given a sequence of tokens \(x_{1:T} = (x_1, x_2, \dots, x_T)\), the primary objective is to minimize the cross-entropy loss:
\begin{equation}
    \mathcal{L}_{\text{NTP}} = -\sum_{t=1}^{T-1} \log P_\theta(x_{t+1} \mid x_{1:t})
\end{equation}
where \(x_{1:t}\) denotes the context up to step \(t\), and \(P_\theta(x_{t+1} \mid x_{1:t})\) represents the model's predicted probability distribution over the vocabulary for the next token. 
During inference, NTP operates in a strictly autoregressive manner. The model generates a single token at a time by sampling from the predicted distribution.

\subsection{Multi-Token Prediction (MTP)}
MTP extends the standard NTP paradigm by training the
model to predict $n$ future tokens $x_{t+1:t+n}$ at each position, using a shared
backbone together with one main head and $n-1$ auxiliary heads.
\begin{equation}
    \mathcal{L}_{\text{MTP}} = -\sum_{t=1}^{T-n} \log P_\theta(x_{t+1:t+n} \mid x_{1:t}).
\end{equation}

In recent studies, the overall architecture $\theta$ is decoupled into a shared
backbone with parameters $\theta'$, a main language modeling head $\theta_0$, and
$n-1$ auxiliary output heads with parameters $\{\theta_j\}_{j=1}^{n-1}$. The shared
backbone maps the context $x_{1:t}$ to a hidden representation
$z_{1:t}=f_{\theta'}(x_{1:t})$. The main head reuses $z_{1:t}$ to predict the next
token $x_{t+1}$, while the $j$-th auxiliary head predicts the token at offset
$j{+}1$, i.e. $x_{t+j+1}$. All $n$ heads operate in parallel on the same
representation, so that $n$ future tokens are produced within a single forward
pass. Assuming conditional independence of the future tokens given $z_{1:t}$, the
joint distribution of the next $n$ tokens is factorized as
%

\begin{equation}
\resizebox{\columnwidth}{!}{$\displaystyle
P_{\theta}(x_{t+1:t+n}\mid x_{1:t})
= P_{\theta_0}(x_{t+1}\mid z_{1:t})
\prod_{j=1}^{n-1} P_{\theta_j}(x_{t+j+1}\mid z_{1:t}),
$}
\label{eq:mtp_factorization}
\end{equation}
where $z_{1:t}$ is a deterministic function of the context and
$P_{\theta_j}(\cdot\mid z_{1:t})$ denotes the distribution predicted by the
$j$-th auxiliary head. 
In practice, rather than pretraining from scratch, these
auxiliary heads are commonly attached to an already pretrained LLM and trained
via fine-tuning, retrofitting multi-token prediction onto an existing NTP model
at a fraction of the pretraining cost.

During inference, the MTP model leverages its multiple heads to propose subsequent \(n\) candidate tokens simultaneously. To guarantee generation quality, this parallel drafting mechanism is paired with a verification step, effectively forming a self-speculative decoding framework. The model evaluates the proposed candidates in a single forward pass, accepts the valid tokens, and advances the context window accordingly. This approach accelerates inference while strictly preserving the original output distribution of the LLM.

\section{Methodology}
\label{sec:method}

In this section, we detail \textbf{AdaMTP} (Adaptive Multi-Token Prediction), our framework that dynamically adjusts the multi-token prediction horizon based on the intrinsic predictability of the sequence. AdaMTP consists of an entropy-based data segmentation algorithm, a two-stage adaptive training pipeline, and a dual-mode accelerated decoding mechanism for inference.

\subsection{Entropy-Based Data Segmentation}
\label{subsec:data_segmentation}
During training, standard MTP forces the model to predict a fixed number of future tokens at every time step, regardless of the difficulty or predictability of the context. This rigid approach can be highly detrimental: forcing the model to predict across high-entropy boundaries---where future tokens are conceptually unrelated to the current context---injects noisy, conflicting training signals. Since the auxiliary MTP heads and the main language modeling head share the same underlying hidden representations, these noisy gradients can severely interfere with the main head's optimization, ultimately degrading the base model's core capabilities. 

To alleviate this optimization burden and prevent representation interference, we propose an entropy-based segmentation algorithm that groups tokens according to their predictive uncertainty. Given a sequence of tokens $X = (x_1, x_2, \dots, x_N)$, we use the exact same pre-trained base model designated for supervised fine-tuning (SFT) as our reference model to compute the entropy of the next-token distribution at each position $t$:
\begin{equation}
\resizebox{0.89\columnwidth}{!}{$\displaystyle
E_t = \mathcal{H}(P(\cdot \mid x_{<t})) = - \sum_{v \in \mathcal{V}} P(v \mid x_{<t}) \log P(v \mid x_{<t})$}\,,
\end{equation}
where $\mathcal{V}$ is the vocabulary. Using the SFT base model as the reference ensures that the computed uncertainty perfectly aligns with the target model's intrinsic probability distribution. We then calculate the delta entropy between consecutive tokens as $\Delta E_t = E_{t+1} - E_t$. Conceptually, as the model generates tokens within a cohesive semantic chunk, the local context becomes increasingly constrained, causing the predictive entropy to exhibit a roughly monotonically decreasing trend (i.e., uncertainty tends to diminish as the phrase nears completion). Conversely, a large positive $\Delta E_t$ signifies a sudden spike in uncertainty, typically corresponding to crossing a linguistic boundary or transitioning to a new semantic unit. This entropy dynamic, illustrated in Figure~\ref{fig:intro}, implies that detecting these surges alone suffices to cleanly recover coherent semantic groups.

Therefore, we segment the sequence into contiguous groups by splitting at indices where the surge in entropy exceeds a predefined threshold ($\Delta E_t > \tau$). The threshold $\tau$ is calibrated via a dataset-level search so that the resulting average group size matches the total number of heads $n$ (i.e., the maximum prediction depth), thereby aligning the segmentation granularity with the model's multi-token prediction capacity.
Let $G_k = [s_k, e_k)$ denote the $k$-th token group spanning from index $s_k$ to $e_k - 1$. For a token $x_t$ located within $G_k$, we define its adaptive prediction depth $d_t$ as follows:
\begin{equation}
    d_t = 
    \begin{cases} 
      e_k - t - 1, & \text{if } t < e_k - 1 \\
      |G_{k+1}|, & \text{if } t = e_k - 1 
    \end{cases}
\end{equation}
where $|G_{k+1}| = e_{k+1} - s_{k+1}$ denotes the length of the subsequent group. Intuitively, tokens strictly inside a predictable group only predict the remainder of their current group, while the boundary token (which precedes a high-entropy transition) is tasked with predicting the entirety of the next group. This design ensures that the training prediction depth flexibly adapts to the natural chunking of language. Throughout, we use \emph{horizon} to denote the number of future tokens the model is asked to predict at a step, and \emph{(adaptive) depth} $d_t$ to denote the per-token training target assigned by our method. Since the model is equipped with only $n-1$ auxiliary heads, the effective supervised depth is capped at the horizon $n$: for any token, whenever $d_t > n$, supervision is applied only up to offset $n$, i.e., $\min(d_t, n)$.



\subsection{AdaMTP Training}
\label{subsec:training}

At the heart of AdaMTP is its adaptive training objective: instead of forcing every token to predict a fixed number of future tokens, we supervise each token only within its adaptive prediction depth $d_t$, masking out any prediction that crosses a semantic boundary. To instill this objective into a pretrained LLM while minimizing disruption to its foundational language modeling ability, we adopt a two-stage pipeline---an auxiliary-head warm-up followed by joint fine-tuning---following the standard training practice for retrofitting MTP heads onto pretrained models \citep{cai2024medusa,liu2025mtp}.

\paragraph{Auxiliary-Head Warm-Up.} 
Our training procedure builds upon a pretrained base language model. To equip the model with multi-token prediction capabilities, we augment the base LLM with \(n-1\) auxiliary output heads, each responsible for predicting a token at a specific future offset. Architecturally, each auxiliary head is implemented as a multi-layer perceptron (MLP) followed by a linear projection layer that maps the hidden states directly into the vocabulary space. The primary objective of this initial stage is to align these newly introduced heads with the base LLM. To achieve this, we freeze the LLM backbone and its original language modeling head, restricting parameter optimization entirely to the auxiliary heads. During this stage, we utilize self-distilled data for training, which is generated by feeding prompts into the base LLM and collecting its outputs. The training objective is formulated as:
\begin{equation}
    \mathcal{L}_{\text{warm-up}} = \sum_{j=1}^{n-1} \sum_{t=1}^{T-j-1} \mathcal{L}_{\text{CE}}\left(\text{Head}_j(\mathbf{h}_t), x_{t+j+1}\right),
\end{equation}
where $\mathbf{h}_t$ is the frozen hidden state of the backbone at time step $t$, and $\mathcal{L}_{\text{CE}}$ denotes the standard cross-entropy loss. 

\paragraph{Adaptive Joint Training.}
In the second stage, we conduct joint training by applying Low-Rank Adaptation (LoRA) across all model components, encompassing both the LLM backbone and the output heads. During this joint training process, we utilize the pre-computed adaptive depths \(d_t\) to selectively mask the MTP loss. 

Let \(\mathbf{h}_t\) denote the hidden state at time step \(t\). The base model predicts the next token \(x_{t+1}\), while the \(j\)-th MTP head (\(1 \le j < n\)) predicts the future token \(x_{t+j+1}\). Unlike traditional MTP training, which uniformly penalizes predictions up to a fixed depth regardless of context difficulty, our approach dynamically masks the loss for tokens beyond the adaptive depth \(d_t\). Consequently, the adaptive MTP loss is defined as:
\begin{equation}
\begin{split}
    \mathcal{L}_{\text{MTP}} = \sum_{j=1}^{n-1} \sum_{t=1}^{T-j-1} & \mathbb{I}(j + 1 \le d_t) \\
    & \cdot \mathcal{L}_{\text{CE}}\left(\text{Head}_j(\mathbf{h}_t), x_{t+j+1}\right),
\end{split}
\end{equation}
where \(\mathbb{I}(\cdot)\) is the indicator function and \(\mathcal{L}_{\text{CE}}\) is the cross-entropy loss. To enable joint optimization, the total loss is formulated as a weighted sum of the standard language modeling loss \(\mathcal{L}_{\text{LM}}\) (for the base model) and the adaptive MTP loss \(\mathcal{L}_{\text{MTP}}\):
\begin{equation}
    \mathcal{L}_{\text{total}} = \mathcal{L}_{\text{LM}} + \lambda \mathcal{L}_{\text{MTP}},
\end{equation}
where \(\lambda\) controls the contribution of the auxiliary heads. This adaptive joint training prevents the model from attempting to predict excessively far into unpredictable futures, thereby concentrating its capacity on highly certain and structured predictions.

\subsection{Inference Procedure}
\label{subsec:inference}
During inference, the trained MTP heads are used for self-speculative decoding: at each step they draft several candidate future tokens, and the base model verifies these candidates in a single forward pass, accepting only the prefix consistent with its own predictions. Because every accepted token is validated by the base model, this draft-then-verify procedure is lossless---it yields exactly the same output as standard autoregressive decoding while reducing the number of sequential steps and thus accelerating generation. Within this framework, we provide two generation settings:

\paragraph{Fixed-Horizon Generation.} 
In this default setting, the MTP heads always draft the full horizon of $n$ candidate tokens at every step. Following Medusa \citep{cai2024medusa}, these candidates are organized into a token tree and verified by the base model in a single forward pass. Since this verification pass is dominated by the base model's forward computation and, at small batch sizes, is only weakly sensitive to the number of candidates it checks, drafting the maximum number of candidates maximizes the expected number of tokens accepted per step.

\paragraph{Adaptive-Horizon Generation.}
Alternatively, we offer an adaptive strategy that dynamically determines how many candidates to generate from the real-time entropy delta: the MTP heads stop drafting further into the future once the entropy increase between consecutive predicted tokens exceeds a threshold. This prunes the candidate tree and reduces the number of tokens the base model must verify. Since it only forgoes low-confidence tail candidates that would largely have been rejected anyway, under an appropriate threshold it incurs no loss in per-step acceptance length relative to fixed-horizon generation. The resulting saving is marginal for single-sample decoding---where verifying a few extra candidates adds negligible latency---but becomes substantial under large-batch, compute-bound serving, whose latency scales with the total number of candidates verified.
\begin{table*}[t]
\centering
\renewcommand{\arraystretch}{1.25}       
\resizebox{\textwidth}{!}{%
\begin{tabular}{cl|cc|cccc|cc|c}
\toprule
& & Math500 & GSM8K & MBPP & MBPP$^+$ & HumanEval & HumanEval$^+$ & MMLU & IFEval & Avg. \\
\hline
& \textcolor{gray!45}{Base} & \textcolor{gray!45}{3.20} & \textcolor{gray!45}{9.02} & \textcolor{gray!45}{62.43} & \textcolor{gray!45}{52.12} & \textcolor{gray!45}{37.80} & \textcolor{gray!45}{30.49} & \textcolor{gray!45}{63.45} & \textcolor{gray!45}{16.91} & \textcolor{gray!45}{34.43} \\
& NTP & 5.60 & 11.30 & 61.34 & 50.95 & 42.26 & 35.37 & 63.64 & 20.08 &36.32  \\
& MTP & 5.00 & 11.30 & 60.38 & 50.00 & 41.46 & 35.98 & 63.33 & 19.74 &35.90  \\
\cline{2-11}
\rowcolor{adamtpblue} \multirow{-4}{*}{\cellcolor{white}\rotatebox[origin=c]{90}{\textit{Llama3.1-8B}}} 
& AdaMTP & 7.20 & 13.12 & 61.64 & 50.26 & 42.68 & 35.98 & 63.67 & 20.26 & \textbf{36.85} \\
\midrule
& \textcolor{gray!45}{Base} & \textcolor{gray!45}{62.80} & \textcolor{gray!45}{54.85} & \textcolor{gray!45}{75.20} & \textcolor{gray!45}{64.06} & \textcolor{gray!45}{78.05} & \textcolor{gray!45}{71.20} & \textcolor{gray!45}{71.76} & \textcolor{gray!45}{41.65} & \textcolor{gray!45}{64.95} \\
& NTP &49.20  &52.75  &76.50  &64.29  &77.44  &69.26  &71.58  &42.35  &62.92  \\
& MTP & 48.60 & 47.69 & 74.34 & 63.76 & 76.22 & 70.12 & 71.77 & 40.29 &61.60  \\
\cline{2-11}
\rowcolor{adamtpblue} \multirow{-4}{*}{\cellcolor{white}\rotatebox[origin=c]{90}{\textit{Qwen2.5-7B}}} 
& AdaMTP & 49.40 & 50.72 & 76.19 & 64.81 & 78.05 & 71.95 & 71.79 & 42.57 & \textbf{63.19} \\
\midrule
& \textcolor{gray!45}{Base} & \textcolor{gray!45}{0.00} & \textcolor{gray!45}{9.76} & \textcolor{gray!45}{72.22} & \textcolor{gray!45}{58.99} & \textcolor{gray!45}{45.12} & \textcolor{gray!45}{35.37} & \textcolor{gray!45}{24.39} & \textcolor{gray!45}{29.52} & \textcolor{gray!45}{34.42} \\
& NTP &9.00  &9.33  &70.63  &58.99  &61.02  &56.10  &70.73  &29.94  &45.72  \\
& MTP &12.60  &13.86  &65.34  &54.50  &59.15  &51.22  &71.82  &31.41  &44.99  \\
\cline{2-11}
\rowcolor{adamtpblue} \multirow{-4}{*}{\cellcolor{white}\rotatebox[origin=c]{90}{\textit{Gemma3-12B}}} 
& AdaMTP &16.00  &14.81  &66.08  &54.50  &60.37  &53.05  &72.04  &32.13  & \textbf{46.12} \\
\bottomrule
\end{tabular}%
} 
\caption{Performance comparison among different prediction paradigms (NTP, MTP, and our AdaMTP) across diverse tasks and benchmarks. For each backbone, the best average result (Avg.) among NTP, MTP, and AdaMTP is highlighted in bold.}
\label{tab:performance_comparison}
\end{table*}

\section{Experiments}
\subsection{Experimental Settings}
\begin{list}{}{\leftmargin=0em}
\item{\textbf{Datasets.}}
In alignment with the previous work \cite{liu2025mtp}, our training corpus is assembled from the Math \cite{hendrycks2021measuring}, Evol-Instruct-Code \cite{luo2023wizardcoder, chaudhary2023code}, and Alpaca-GPT4 \cite{peng2023instruction} datasets. The training procedure is divided into two phases. Initially, the entire dataset is leveraged to perform self-distillation. Subsequently, the second phase utilizes a randomly sampled subset of 10,000 instances, distributed across mathematical, programming, and general domains in a 4:4:2 ratio. To rigorously assess the proposed methodologies, we employ a diverse suite of benchmarks: Math500 \cite{lightman2023let} and GSM8K \cite{cobbe2021training} (both evaluated in a 4-shot setting) for mathematical reasoning; MBPP, MBPP+ \cite{austin2021program, liu2023your}, HumanEval, and HumanEval+ \cite{chen2021evaluating, liu2023your} for code generation capabilities; alongside MMLU \cite{hendrycks2020measuring} and IFEval \cite{zhou2023instruction} to measure general proficiency.


\item{\textbf{Evaluation Metrics.}}
To evaluate task performance, we report accuracy for mathematical and general-domain benchmarks, while employing the pass@1 metric for code generation tasks. Furthermore, we assess efficiency via a speedup ratio relative to standard autoregressive NTP decoding on the same model.
\item{\textbf{Base LLMs.}}
Our experimental framework employs three base large language models: Llama-3.1 (8B), Qwen-2.5 (7B), and Gemma-3 (12B). This selection was made to ensure a comprehensive evaluation across a diverse spectrum of model architectures and parameter capacities.
\item{\textbf{Baselines.}}
To evaluate generation efficacy, we benchmark our approach against two established paradigms: standard NTP and MTP. These baselines serve to measure the models' capacity to produce accurate and contextually appropriate text. For the efficiency analysis, standard autoregressive NTP decoding provides the $1\times$ reference against which speedups are measured, while we compare AdaMTP with standard MTP, both of which employ self-speculative decoding for lossless acceleration.

\item{\textbf{Implementation Details.}}
During the initial head warm-up stage, we freeze the LLM backbone and exclusively train the prediction heads for 1 epoch with a learning rate of \(1 \times 10^{-3}\). In the subsequent stage, we utilize LoRA (rank \(r = 32\), \(\alpha = 16\)) to fine-tune the model for 3 epochs with a learning rate of \(1 \times 10^{-5}\). We set the prediction depth to $n = 4$ and the auxiliary MTP loss weight to $\lambda = 0.1$ by default. To ensure a fair comparison, identical training configurations are applied to the standard MTP baseline. All experiments are conducted on four NVIDIA H800 GPUs with a total batch size of 256.

\end{list}

\subsection{Overall Performance}

Table~\ref{tab:performance_comparison} compares AdaMTP against NTP and standard MTP across three base models and eight benchmarks; we highlight the key findings below.


\paragraph{AdaMTP consistently achieves the best overall performance.}
Across all three backbones, AdaMTP attains the highest average score---\textbf{36.85} on Llama-3.1-8B, \textbf{63.19} on Qwen-2.5-7B, and \textbf{46.12} on Gemma-3-12B---surpassing both NTP and standard MTP. This consistency across model families and scales shows that adaptively aligning the prediction horizon with sequence predictability is a robust, architecture-agnostic strategy. Intriguingly, on Qwen-2.5-7B all fine-tuning paradigms (including NTP) fall below the Base model, an observation consistent with prior findings~\citep{liu2025mtp}; we attribute this to the distribution shift of our fine-tuning corpus rather than the MTP training paradigm, as plain NTP is affected identically. Even in this regime AdaMTP still outperforms both NTP and standard MTP, indicating that the adaptive objective remains beneficial regardless of data quality. As our focus is isolating the effect of the adaptive horizon relative to standard MTP, we leave higher-quality data curation to future work.

\begin{table}[t]
\centering
\renewcommand{\arraystretch}{1.1}
\begin{tabular*}{\columnwidth}{@{\extracolsep{\fill}} c l c c c @{}}
\toprule
& & GSM8K & HumanEval & IFEval \\
\midrule
\multirow{3}{*}{\rotatebox{90}{\footnotesize Llama3.1}}
  & NTP    & 1.00$\times$ & 1.00$\times$ & 1.00$\times$ \\
  & MTP    & 1.65$\times$ & 1.86$\times$ & 1.48$\times$ \\
  & \textbf{AdaMTP} & \textbf{2.12$\times$} & \textbf{2.01$\times$} & \textbf{1.52$\times$} \\
\midrule
\multirow{3}{*}{\rotatebox{90}{\footnotesize Qwen2.5}}
  & NTP    & 1.00$\times$ & 1.00$\times$ & 1.00$\times$ \\
  & MTP    & 1.84$\times$ & 1.55$\times$ & 1.40$\times$ \\
  & \textbf{AdaMTP} & \textbf{1.87$\times$} & \textbf{1.56$\times$} & \textbf{1.43$\times$} \\
\midrule
\multirow{3}{*}{\rotatebox{90}{\footnotesize Gemma3}}
  & NTP    & 1.00$\times$ & 1.00$\times$ & 1.00$\times$ \\
  & MTP    & 2.38$\times$ & 1.59$\times$ & 1.57$\times$ \\
  & \textbf{AdaMTP} & \textbf{2.75$\times$} & \textbf{1.62$\times$} & \textbf{1.61$\times$} \\
\bottomrule
\end{tabular*}
\caption{Inference speedup comparison of NTP, MTP, and AdaMTP across three backbones.}
\label{tab:speedup}
\end{table}

\paragraph{Standard MTP suffers from representation interference.}
Standard MTP consistently underperforms NTP on average across all three models. This corroborates our hypothesis: forcing the auxiliary heads to predict a rigid number of tokens across high-entropy boundaries injects noisy, conflicting gradients into the shared backbone, corrupting its core capabilities. AdaMTP instead masks these unpredictable predictions, suppressing the interfering signals and exceeding the performance of NTP.
\paragraph{Gains span math reasoning, code, and general tasks.}
AdaMTP improves consistently across task categories, most notably on mathematical reasoning: since these tasks demand long-range planning, concentrating multi-token supervision within cohesive, predictable chunks yields cleaner training signals. On code generation, it recovers the losses incurred by standard MTP and matches or exceeds NTP. It also preserves general-domain ability, achieving the best IFEval and MMLU scores among the trained paradigms.

\subsection{Inference Acceleration}
Beyond task performance, a central promise of the MTP paradigm is its ability
to accelerate decoding via self-speculative generation. Table~\ref{tab:speedup}
reports the speedup ratio of NTP, standard MTP, and AdaMTP on GSM8K,
HumanEval, and IFEval across all three backbones, using autoregressive NTP
decoding as the reference ($1.00\times$); here AdaMTP adopts its default
Fixed-Horizon decoding, the same scheme used by standard MTP. Across every
backbone and benchmark, AdaMTP delivers substantial acceleration over NTP,
ranging from $1.43\times$ up to $2.75\times$, with the largest gains on the
highly structured GSM8K task (e.g., $2.75\times$ on Gemma-3-12B). More
importantly, AdaMTP consistently surpasses standard MTP---for instance, improving
the GSM8K speedup from $1.65\times$ to $2.12\times$ on Llama-3.1-8B---while simultaneously achieving
superior task performance. Because the two share an identical inference
procedure, this acceleration gain over MTP arises purely from our adaptive
training objective, which yields auxiliary heads whose drafts are more
frequently accepted by the verifier, allowing AdaMTP to enhance the inference
speedups that make MTP attractive in practice.

\begin{figure}[t!]
\setlength{\abovecaptionskip}{-0.005mm} 
\setlength{\belowcaptionskip}{-2mm} 
  \centering
\includegraphics[width=0.99\columnwidth]{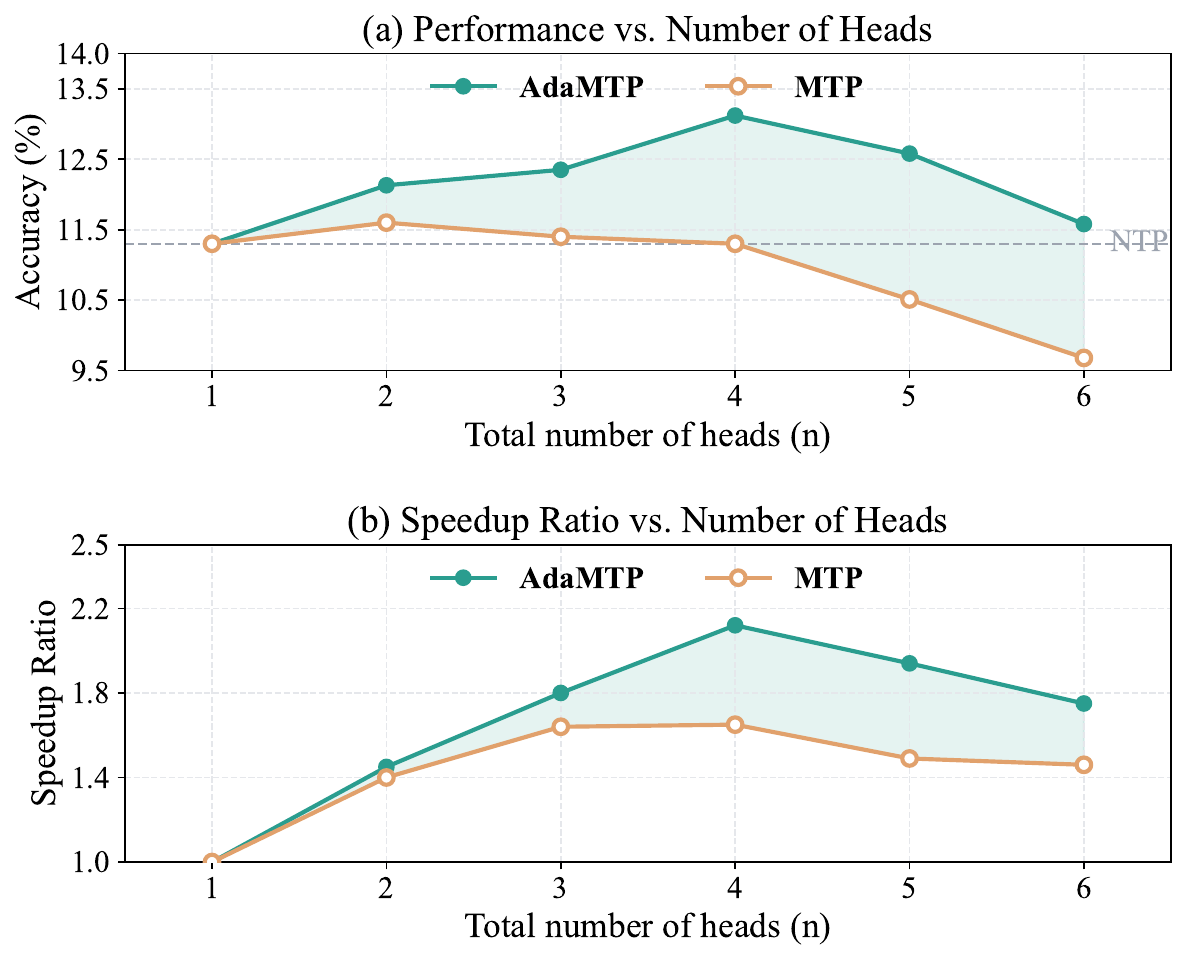}
\caption{Impact of the number of prediction heads $n$ on Llama-3.1-8B
($n=1$ corresponds to NTP). AdaMTP stays consistently above standard MTP.}
\label{fig:heads}
\end{figure}






\subsection{Impact of the Number of Prediction Heads}
We vary the total number of heads from $n{=}1$ (i.e., NTP) to $n{=}6$ on
GSM8K with Llama-3.1-8B, holding all other settings fixed. As shown in
Figure~\ref{fig:heads}(a), the accuracy of standard MTP declines
almost monotonically as $n$ grows, falling from $11.60$ at $n{=}2$ to
$9.68$ at $n{=}6$---well below the $11.30$ NTP reference. This is a direct
consequence of representation interference: each additional head forces
the shared backbone to predict one token further ahead, crossing more
high-entropy semantic boundaries and injecting proportionally more noisy
gradients. AdaMTP instead masks precisely these cross-boundary
predictions, so it stays above NTP throughout and peaks at $n{=}4$ with
$13.12$; accordingly, its gap over MTP widens steadily with $n$ (from
$+0.53$ at $n{=}2$ to $+1.90$ at $n{=}6$). In terms of inference speedup
(Figure~\ref{fig:heads}(b)), AdaMTP likewise dominates standard MTP
at every operating point and degrades more gracefully as $n$ increases, since concentrating
supervision within predictable chunks keeps even its deepest drafts
frequently acceptable to the verifier. Overall, across every head budget
$n$, AdaMTP consistently surpasses standard MTP in both accuracy and
inference speedup, confirming that its advantage is robust to the number
of prediction heads rather than tied to a particular setting.



\subsection{Discussion on Adaptive-Horizon Decoding}

\begin{figure}[t!]
\setlength{\abovecaptionskip}{-0.005mm} 
\setlength{\belowcaptionskip}{-5mm} 
  \centering
\includegraphics[width=0.99\columnwidth]{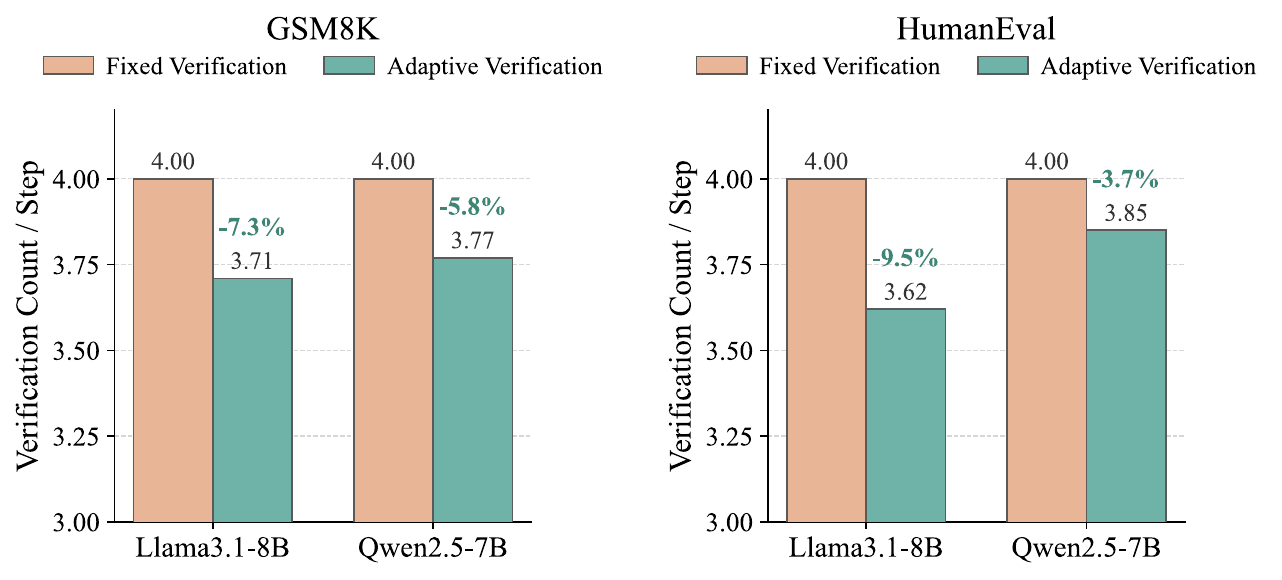}
  \caption{{Average number of candidate tokens verified per decoding step} under the fixed-horizon and adaptive-horizon generation.  The adaptive strategy reduces the per-step verification cost without degrading task accuracy.}
  \label{fig:count}
\end{figure}

We now compare AdaMTP's two inference modes---Fixed-Horizon and Adaptive-Horizon Generation---to quantify the benefit of entropy-based candidate pruning. To this end, we evaluate Adaptive-Horizon against the Fixed-Horizon baseline on GSM8K and HumanEval with Llama3.1-8B and Qwen2.5-7B, measuring the average number of tokens verified per step. As shown in Figure~\ref{fig:count}, the adaptive strategy consistently reduces the verification burden while keeping accuracy statistically indistinguishable from the baseline, confirming that the pruned tail candidates were largely redundant. This reduction, however, yields little speedup for single-sample inference, which is \emph{memory-bandwidth bound}: latency is dominated by streaming the model weights from DRAM, so trimming a few candidates neither reduces weight loading nor shortens the already-parallel verification pass. Under large-batch, \emph{compute-bound} inference, by contrast, latency grows with the total number of candidate tokens, so pruning directly cuts the workload and yields clear throughput gains as batch size increases.

\section{Related Work}

\subsection{Multi-Token Prediction}
\citet{qi2020prophetnet} introduce $n$-step-ahead prediction in sequence-to-sequence pretraining to encourage future planning and reduce overfitting to local correlations.
\citet{gloeckle2024better} formalize multi-token prediction by adding parallel heads during pretraining, improving reasoning over next-token prediction, and \citet{basharin2024faster} generalize these independent heads to a rank-$r$ canonical tensor decomposition to better capture dependencies among future tokens.
MTP has also been adopted at scale during the pretraining of industrial LLMs \citep{liu2024deepseek,xiaomi2025mimo} for better data efficiency and long-horizon planning.
Beyond pretraining from scratch, a more efficient line retrofits an existing pretrained NTP model into an MTP architecture \citep{cai2025fastmtp}: Medusa \citep{cai2024medusa} attaches multiple lightweight heads that each forecast a token at a distinct future offset at a fraction of the pretraining cost; \citet{samragh2025your} equip the model with gated LoRA modules and a learnable sampler for simultaneous multi-token prediction; and L-MTP \citep{liu2025mtp} adds a leap-based mechanism that predicts non-sequential positions in a single forward pass to capture longer-range dependencies.
These methods collectively show that retrofitting a pretrained NTP model with extra prediction heads is a practical, cost-effective route to multi-token prediction, yielding substantial inference speedups while preserving generation quality.
In addition, some approaches use MTP purely as an auxiliary training objective to improve the generation quality of NTP models, and thus provide no inference acceleration, since the model still decodes one token at a time at inference. For instance, MuToR \citep{gerontopoulos2025multi} interleaves learnable register tokens into the training sequence to predict future targets, while TOP \citep{zuhri2025predicting} replaces exact future-token prediction with a learning-to-rank loss that orders upcoming tokens by proximity.

\vspace{-0.5em}
\subsection{LLM Inference Acceleration}
The growing cost of LLM inference has motivated acceleration methods that target different bottlenecks. One line reduces the per-step cost via model compression---quantization \citep{hubara2018quantized,kim2023squeezellm,lin2024awq}, pruning \citep{frantar2023sparsegpt,sun2023simple,ma2023llm,gao2024disp}, and knowledge distillation \citep{gu2023minillm,hinton2015distilling,hsieh2023distilling,ho2023large}---and efficient attention, whether linear \citep{katharopoulos2020transformers,yang2024parallelizing}, sparse \citep{child2019generating,lu2025moba}, or low-rank \citep{liu2024deepseek}. A second line improves system-level throughput through operator fusion \citep{dao2022flashattention,dao2023flashattention}, KV-cache management \citep{kwon2023efficient,zheng2024sglang}, and parallelism \citep{nvidia2023tensorrtllm}. Orthogonally, a third line reduces the number of sequential decoding steps: speculative decoding \citep{leviathan2023fast,chen2023accelerating,miao2024specinfer} losslessly amortizes autoregressive generation by cheaply drafting candidate tokens and validating them in a single target-model forward pass, with candidates proposed either by a separate lightweight model \citep{leviathan2023fast,yang2025multi,zhou2023distillspec} or by the target model itself via auxiliary prediction heads \citep{stern2018blockwise,cai2024medusa,li2024eagle}.


\vspace{-0.5em}
\subsection{Adaptive Modeling}
Motivated by the non-uniform information density of natural language, recent work dynamically adjusts processing granularity rather than treating all tokens uniformly \citep{barrault2024large}. The two most relevant to ours both adapt granularity on the \emph{input or representation} side---the Byte Latent Transformer \citep{pagnoni2025byte} patches bytes for the encoder, and Dynamic Large Concept Models \citep{qu2025dynamic} compress tokens into concepts---yet still emit a single token (or byte) per step and therefore cannot accelerate decoding. AdaMTP instead applies entropy-based, variable-length segmentation to the \emph{multi-token prediction} itself: it turns each segment into an adaptive prediction depth that masks cross-boundary supervision, simultaneously protecting the pretrained backbone and enabling accelerated multi-token generation.
\section{Conclusion}
In this paper, we identified a key limitation of existing MTP: its fixed-length horizon ignores the non-uniform information density of language, forcing auxiliary heads to predict across high-entropy boundaries and injecting noisy gradients that degrade the shared backbone. To address this, we proposed AdaMTP, which uses entropy-based segmentation to assign each token an adaptive prediction depth and applies a dynamically masked MTP objective that suppresses loss for cross-boundary predictions. Across mathematical reasoning, code, and general benchmarks on three backbones, AdaMTP consistently surpasses NTP and standard MTP in both task performance and inference speedup, showing that aligning the multi-token objective with local predictability is a robust, architecture-agnostic way to retrofit pretrained LLMs with efficient multi-token generation. Promising future directions include higher-quality data curation and stronger adaptive-horizon pruning for high-throughput batched inference.



\bibliography{reference}

\end{document}